\documentclass[a4paper,fleqn]{cas-dc}

\usepackage[authoryear]{natbib}

\def\tsc#1{\csdef{#1}{\textsc{\lowercase{#1}}\xspace}}
\tsc{WGM}
\tsc{QE}
\tsc{EP}
\tsc{PMS}
\tsc{BEC}
\tsc{DE}

\begin{document}
\let\WriteBookmarks\relax
\def\floatpagepagefraction{1}
\def\textpagefraction{.001}
\shorttitle{Pattern Recognition Letters}
\shortauthors{Namkyung et~al.}
\title[mode=title]{Beyond Learning: A Training-Free Alternative to Model Adaptation}
\author[1]{Namkyung Yoon\fnref{fn1}}
\ead{nkyoon93@korea.ac.kr}

\author[1]{Kyeonghyun Yoo}
\ead{seven1705@korea.ac.kr}

\author[1]{Wooyong Jung}
\ead{jy17347@korea.ac.kr}

\author[1]{Sanghong Kim}
\ead{sanghongkim@korea.ac.kr}

\author[1]{Hwangnam Kim\corref{cor1}\fnref{fn1}}
\ead{hnkim@korea.ac.kr}

\fntext[fn1]{N.\ Yoon and H.\ Kim contributed equally to this work.}

\affiliation[1]{
  organization={School of Electrical Engineering, Korea University},
  city={Seoul},
  postcode={02841},
  country={Republic of Korea}
}

\cortext[cor1]{Corresponding author}

\begin{abstract}
Despite the continuous research and evolution of language models, they sometimes underperform previous versions. Existing approaches to overcome these challenges are resource-intensive, highlighting the need for alternatives that enable immediate action.
We assume that each language model has a local module inside that is suitable for a specific function.
First, this work identifies a set of modules showing consistent and local activation changes under an inference workload through activation-based analysis.
Subsequently, we transplant an internal module that is properly activated for a specific task into the target model, leading to immediate and measurable functional changes without additional training or fine-tuning.
To experimentally demonstrate the effectiveness of the transplant technique, we quantify the relationship between transplant strength and performance improvement under different conditions for two language models.
In the cross-generation setting, we find that transplanting activation-selected modules can substantially improve the underperforming model, reaching up to twice the target baseline and achieving gap-based recovery above $100\%$.
Moreover, in transplant experiments between a base model and its instruction-tuned counterpart, transplantation improves the underperforming model toward the stronger baseline, yielding up to about 2.33 times the target baseline with gap-based recovery reaching up to $100\%$ in the best case.
These results show that meaningful capacity transfer can be realized through the implantation of highly localized modules implied by language models.
Overall, this work provides empirical evidence for task-localized modularity in language models and presents a new research area: model transplantation.
\end{abstract}


\begin{graphicalabstract}
\includegraphics[width=\linewidth]{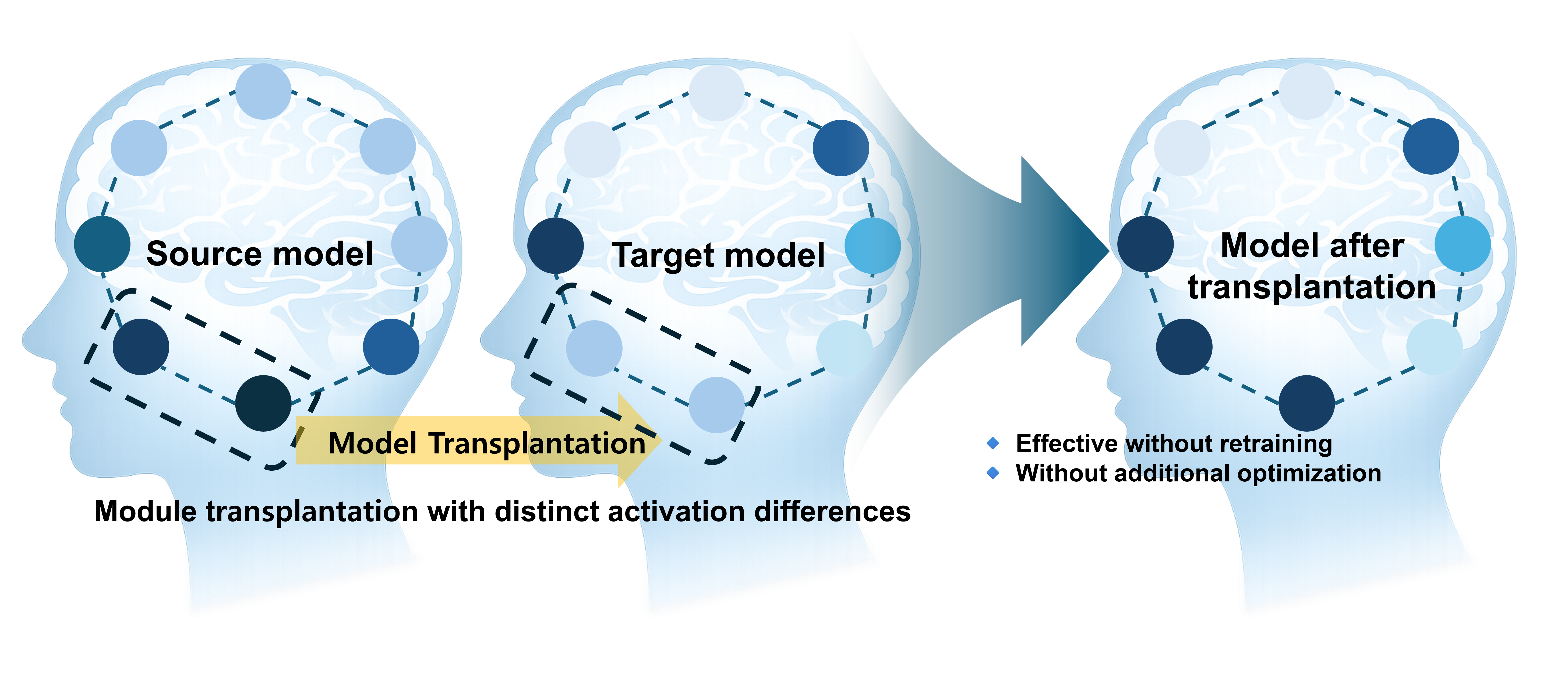}
\end{graphicalabstract}

\begin{highlights}
\item We propose a transplant technique for improving language models without training.
\item We propose a compatibility diagnostic for model transplantation. 
\item We validate the transplant technique across different conditions and models. 
\item We analyze optimal transplant conditions with a new metric.
\end{highlights}


\maketitle
\section{Introduction}

Recent advances in artificial intelligence and computational infrastructure have led to the emergence of language models with strong reasoning ability and domain-specific expertise \cite{kumar2024large}.
Despite these advances, such capabilities remain largely confined within the fixed parameterization of each individual model, limiting the direct reuse or transfer of learned behaviors across models \cite{hadi2023large, ling2023domain}.
As a result, adapting a model to a new domain typically requires additional training procedures, such as fine-tuning, parameter-efficient adaptation, model merging, or knowledge distillation, all of which modify parameters through optimization or statistical combination \cite{han2024parameter, wang2025parameter, fang2025knowledge, yang2024model, liu2023pre}.

While these approaches have proven effective, they do not provide a mechanism for directly transferring internal functional components between models.
In contrast to biological systems, where functional recovery or augmentation can be achieved through localized transplantation \cite{murray2002transplantation}, existing model adaptation techniques provide limited interpretability regarding where and how new features are integrated \cite{sun2024bbox}.
This gap raises fundamental questions about whether pre-trained language models contain internal components that can be functionally communicated in a local and direct manner.

In this work, we aim to address this gap by investigating the internal behavior of language models during inference, using activation-based analysis of their constituent components.
We hypothesize that a small subset of modules within the language model exhibits localized activation changes that are specific to particular domain tasks.

Inspired by these observations, we test whether functional behavior can be reproduced without training by selectively replacing a limited number of layers identified via activation-based analysis.
We refer to this training-free adaptation as model transplantation, in which a small subset of internal components from a source model is directly transplanted into a structurally compatible target model, where corresponding modules match in role and dimensionality.



As a result, the transplanted model recovers task-relevant performance while retaining most of the existing functionality of the target model.
This result suggests that, when maintaining architectural compatibility, the relevant modules for specific domain tasks can be recombined within a language model to improve performance.
In addition, transplantation offers a potential mechanism for rapid rollback when post-training updates inadvertently degrade task-specific behavior. By restoring a small set of functionally relevant modules from a reference model, the degraded performance can be recovered without additional optimization. 

Collectively, these observations provide evidence of module-level portability across models.
They also suggest task-dependent activation structures, analogous to functional specialization in biological systems \cite{ellis2022structural}.


The rest of this paper is organized as follows. Section 2 reviews existing model adaptation techniques, such as fine-tuning. Section 3 describes the proposed transplant technique between language models. Section 4 presents the results of the experiments and evaluations. Section 5 discusses the conclusions.

\begin{figure*}[]
    \centering
    \includegraphics[width=1.0\linewidth]{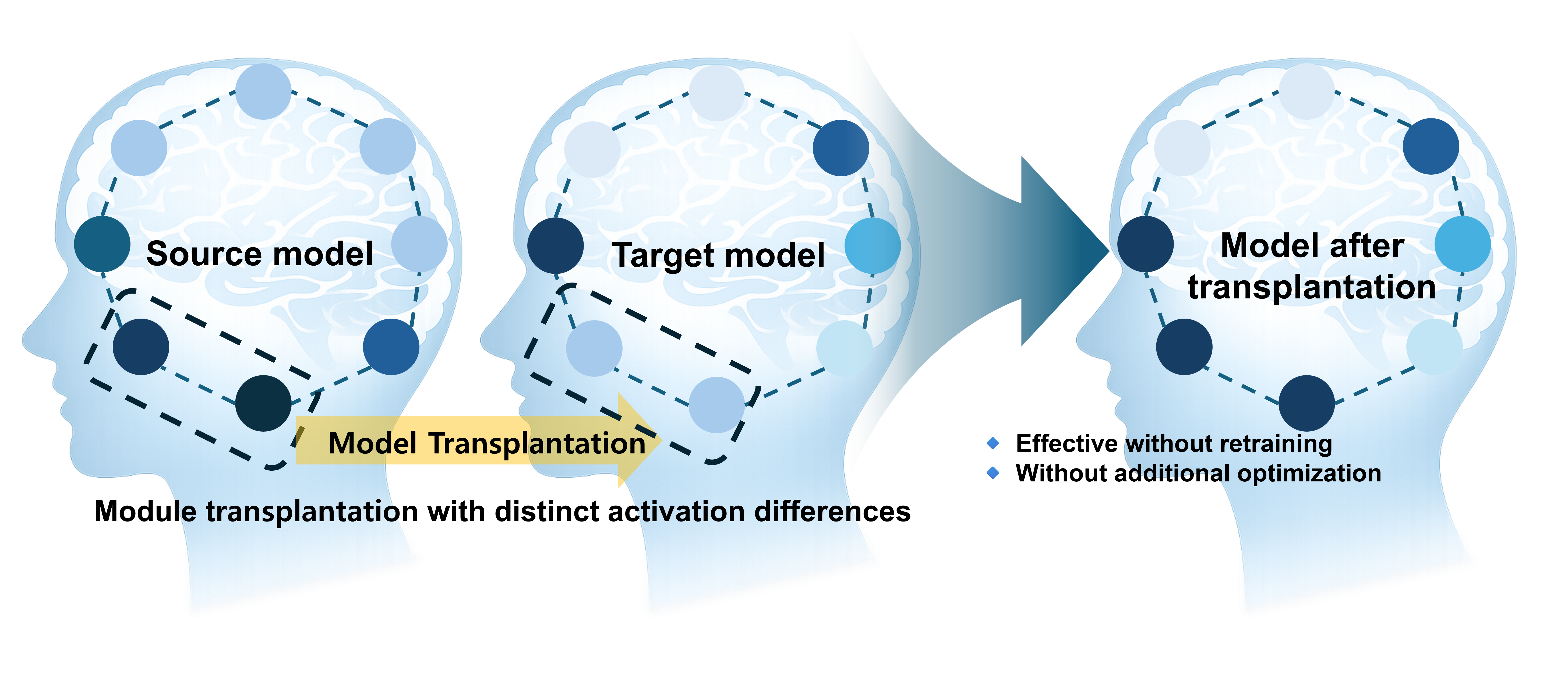}
    \caption{\textbf{Illustration of module-level transplantability in language models.}}
    \label{overall_grafting}
\end{figure*}
\section{Preliminary}

\subsection{Optimization-Based Model Adaptation}

Existing approaches for adapting or combining language models largely rely on optimization-based mechanisms.
Parameter-efficient fine-tuning methods update a small subset of parameters, such as low-rank adapters or selected layers \cite{wang2025parameter, hu2022lora}, through gradient-based optimization using task-specific data.
Model merging techniques combine multiple trained models through parameter averaging or geometric alignment, producing a compromise model that blends behaviors across sources \cite{yang2024model}.
Knowledge distillation transfers knowledge by training a student model to approximate the outputs or representations of a teacher model, resulting in newly learned parameters \cite{fang2025knowledge}.

Despite their differences, these approaches share a common characteristic: model adaptation is achieved through additional training-based optimization.
As a result, they do not directly reuse internal computations via weight-level module reuse, nor do they provide a mechanism to isolate and transfer specific layers responsible for task-specific functionality.

In contrast, the approach studied in this work performs \textit{transplantation} without optimization or retraining.
By selectively replacing structurally compatible layers identified through consistent activation discrepancies, the proposed method enables immediate functional transfer through direct parameter reuse rather than learned approximation.

\subsection{Motivation}
Language models are continuously studied and updated across generations to improve overall performance and generalization.
However, in practice, such updates can introduce performance regressions, whereby the new model performs worse than the previous version on tasks that were previously handled reliably.
This can lead to critical outcomes, as task-specific reliability may degrade in real-world deployments despite apparent improvements in aggregate benchmark metrics.
Resolving these performance regressions through traditional adaptation techniques usually requires additional training, task-specific data, or architectural modifications.
These requirements impose significant latency and overhead, making them unsuitable for rapid recovery or real-time intervention.

From a system perspective, an ideal solution should be able to selectively restore lost features while preserving most of the updated model behavior without retraining or global parameter modification.
This requirement implicitly assumes that task-related computations are not uniformly distributed across the model, but are concentrated in a subset of inner layers that are disproportionately engaged depending on the task.
Based on these assumptions, it becomes possible to identify and selectively replace only the components that contribute to the degraded behavior.

Building on this motivation, this work investigates transplantation as a method for selective functional recovery and structured model editing in evolving language model systems.

\section{Model Transplant}
The proposed model transplantation is a mechanism that changes operational behavior by replacing layers based on local activation scores as shown in Fig.~\ref{overall_grafting}

\subsection{Pre-Transplant Diagnosis}

Our transplant procedure begins with identifying structurally compatible modules across models.

In this paper, $\mathcal{M}_S$ and $\mathcal{M}_T$ are represented as the source and target models for transplantation.
For a model $\mathcal{M}$, we define the set of candidate modules as follows:
\begin{equation}
\mathcal{L}(\mathcal{M}) = \{\, \ell \mid \ell \text{ is a leaf } \texttt{nn.Linear} \text{ module in } \mathcal{M} \,\},
\end{equation}
where non-transferable components such as embedding layers, normalization layers, dropout layers, and output heads are excluded.

Each module $\ell \in \mathcal{L}(\mathcal{M})$ is represented by its normalized name $n_\ell$ and weight matrix $\mathbf{W}_\ell \in \mathbb{R}^{d_{\text{out}} \times d_{\text{in}}}$.
A module pair $(\ell_S, \ell_T)$ with $\ell_S \in \mathcal{L}(\mathcal{M}_S)$ and $\ell_T \in \mathcal{L}(\mathcal{M}_T)$ is considered compatible if
\begin{equation}
n_{\ell_S} = n_{\ell_T}
\quad \text{and} \quad
\text{shape}(\mathbf{W}_{\ell_S}) = \text{shape}(\mathbf{W}_{\ell_T}).
\end{equation}

The set of transplantable modules is thus defined as follows:
\begin{equation}
\mathcal{C} =
\{ (\ell_S, \ell_T) \mid
n_{\ell_S} = n_{\ell_T},
\ \text{shape}(\mathbf{W}_{\ell_S}) = \text{shape}(\mathbf{W}_{\ell_T}) \},
\end{equation}
with transplantation implemented by direct weight copy as follows:
\begin{equation}
\mathbf{W}_{\ell_T} \leftarrow \mathbf{W}_{\ell_S}, \quad \forall (\ell_S, \ell_T) \in \mathcal{S},
\end{equation}
where $\mathcal{S}$ is a selected subset of compatible module pairs and is defined in the next section based on activation discrepancies.
When present, the corresponding bias terms are copied in the same manner.

All subsequent transplantation experiments operate only on module pairs drawn from $\mathcal{C}$, and we always select $\mathcal{S} \subseteq \mathcal{C}$.
No fine-tuning, architectural modification, or auxiliary adaptation is applied after transplantation, ensuring that observed performance changes reflect functional transfer rather than structural effects.

In practice, we extract only standalone linear modules, including attention projections and feed-forward layers, and normalize their namespaces to ensure one-to-one alignment between models.


\subsection{Selection Criterion for Transplanting}

To quantitatively identify transplant candidates, we compare the activations of the source and target models using a layer-wise activation evaluation (LAE) based on forward-only inference.
The LAE records activation levels across both models during autoregressive inference on input sequences to address specific domains.
This aims to distinguish the local activation differences between the target model to be transplanted and the source model to which the transplant module is donated.

For linear modules $m_i$, such as attention projections or feedforward layers, the activation response at decoding step $t$ is defined as follows:
\begin{equation}
a_{m_i}(t) = \frac{1}{d}\|h_{m_i}(t)\|_1,
\end{equation}
where $h_{m_i}(t)\in \mathbb{R}^d$ represents the post-linear hidden state of the last generated token at step $t$.
The activation mismatch between the source and target models is measured with LAE score, which is defined as:
\begin{equation}
s(m_i)=\mathbb{E}_{x,\ t\le T}\big[|a^{*}_{m_i}(t)-a_{m_i}(t)|\big],
\end{equation}
where $a^{*}_{m_i}(t)$ denotes the activation of the source model and $a_{m_i}(t)$ denotes the activation of the target model under the same input $x$.

This metric provides a stable and interpretable measure of functional divergence by using last-token activations and averaging across decoding steps, capturing persistent task-induced differences while avoiding transient fluctuations.

The layers are ranked according to $s(m_i)$, and the top $K$ layers with the greatest discrepancy are selected as candidates for transplantation.
The transplant is then applied only to this subset and modified locally while retaining most original structures, changing only some of the parameters of the target model.



\subsection{Model Transplantation Procedure}

This section describes the model transplantation procedure, including the diagnosis of architectural compatibility between the source and target models and the criteria for selecting modules for transplantation.

Model transplantation replaces the parameters of a target model with the corresponding parameters from a source model, restricted to structurally aligned linear modules with identical shapes.

In this paper, we define $\theta$ as the target model parameters, $\theta^{*}$ as the source model parameters, and $\mathcal{S}$ as a selected set of transplantable module pairs.
We then select a subset $\mathcal{S} \subseteq \mathcal{C}$ based on LAE scores, and apply transplantation only to pairs in $\mathcal{S}$.
The transplanted model $\theta'$ is constructed as follows:
\begin{equation}
\theta'_{\ell_T}=
\begin{cases}
\theta^{*}_{\ell_S}, & (\ell_S,\ell_T)\in\mathcal{S},\\
\theta_{\ell_T}, & \text{otherwise}.
\end{cases}
\end{equation}

After this procedure, no additional optimization or fine-tuning is applied to the transplanted model, ensuring that any performance change arises solely from the transplanted modules.
We ensure that for each selected module, the data type is maintained by transplanting weights and biases as follows:
\begin{equation}
\mathbf{W}_{\ell_T} \leftarrow \mathbf{W}_{\ell_S}, \qquad
\mathbf{b}_{\ell_T} \leftarrow \mathbf{b}_{\ell_S}.
\end{equation}
As a result, the transplanted model retains the existing architecture while inheriting the functionality of the source model.

\section{Experiment}

In this section, we present an evaluation designed to assess whether the proposed model transplant technique can effectively induce task-specific competency transfer in language models.

\begin{table*}[t]
\centering
\vspace{0.6em}
\caption{
Comparison of key architectural and training characteristics across Phi and Gemma model variants.
}
\resizebox{1.0\textwidth}{!}{
\begin{tabular}{lcccc}
\hline
\textbf{Component}               
& \textbf{Phi-3-mini-4k} 
& \textbf{Phi-3.5-mini} 
& \textbf{Gemma-2-2B} 
& \textbf{Gemma-2-2B-IT} \\
\hline
Context Length                   
& 4{,}096 tokens         
& 128K tokens           
& 8K tokens
& 8K tokens \\

Vocabulary Size                  
& 32{,}064 tokens        
& 32{,}064 tokens       
& 256K tokens
& 256K tokens \\

Architecture                     
& Decoder-only           
& Decoder-only          
& Decoder-only
& Decoder-only \\

Training Corpus                  
& Curated Phi-3 dataset  
& Expanded Phi-3.5 data 
& Gemma pretraining corpus
& Gemma pretraining corpus + instruction tuning \\
\hline
\end{tabular}
}
\label{model_specs}
\end{table*}

\subsection{Implementation and Dataset Details}

We evaluate our method on a mathematical inference task that is known to induce heterogeneous and depth-dependent activation patterns in large language models \cite{feng2023towards}.
First, we conduct transplantation experiments within the Phi family using the open-source \textsc{Phi-3-mini-4k-command} model.
Using the proposed pre-transplant diagnostic procedure, we identify \textsc{Phi-3.5-mini-command} as a structurally compatible model for transplantation.

Phi-3.5 is an in-generation update that largely preserves the architecture of Phi-3 while extending context length and improving instruction following through additional training \cite{abdin2024phi}.

In addition, we include a Gemma-model setting to isolate the effect of instruction tuning \cite{team2025gemma}. 
Specifically, we perform transplantation between \textsc{Gemma-2-2B} and its instruction-tuned counterpart \textsc{Gemma-2-2B-IT}.
This setting explicitly evaluates how specialization induced by instruction tuning affects module transferability.

The architectural and training characteristics of Phi-3, Phi-3.5, and Gemma models are summarized in Table~\ref{model_specs}.
We design a targeted diagnostic workload based on functional specialization to analyze transferability.
Using the pre-transplant diagnostic technique described in Section~3.1, we find that all model pairs used in our experiments achieve compatibility over the candidate linear modules.

All experiments are conducted on the \texttt{HuggingFaceH4/ MATH-500} benchmark, which consists of 500 problems spanning seven mathematical subjects \cite{math500dataset_card}:
\textit{Algebra}, \textit{Intermediate Algebra}, \textit{Prealgebra}, \textit{Precalculus},
\textit{Geometry}, \textit{Number Theory}, and \textit{Counting \& Probability}.


On these datasets, we extract 50 questions by sampling and use them to evaluate transplantation between models.

At this point, we use the same prompt that requires a single boxed final answer from all models for the reliability of the evaluation:
\begin{quote}\small
\texttt{You are a careful math problem solver. Solve step by step and give ONLY the final answer wrapped in \textbackslash boxed\{...\}.}
\end{quote}

Notably, on the selected mathematical reasoning benchmarks, \textsc{Phi-3-mini-4k-instruct} sometimes exhibits stronger task performance than the newer \textsc{Phi-3.5-mini-instruct}, despite \textsc{Phi-3.5} being an updated variant.
This performance reversal provides a controlled setting in which the newer model is not necessarily superior in task accuracy, allowing us to test whether model transplantation can recover task-relevant behavior by transferring only localized components rather than relying on global fine-tuning or full-model replacement.

For the Gemma setting, we evaluate transplantation between the base model \textsc{Gemma-2-2B} and its instruction-tuned counterpart \textsc{Gemma-2-2B-IT}.
Because their relative performance can vary across decoding lengths and evaluation conditions, we first evaluate both models for each experimental configuration and then perform transplantation from the better-performing model to the weaker model.
We evaluate transplantation performance by varying the decoding length using max\_new\_tokens set to 32, 64, 128, 256, and 512, together with the number of transplanted layers K set to 8, 16, 32, 64, and 128.
Unless otherwise stated, we fix SEED to 42, disable sampling, and set top\_p to 1.0.

Accuracy is defined as the fraction of correct predictions over the evaluation set.
Given the model output $\hat{y}_i$ and the ground-truth answer $g_i$ for the $i$-th problem, accuracy is computed as
\[
\mathrm{Acc}
=
\frac{1}{N}
\sum_{i=1}^{N}
\mathbb{1}\!\left[
\mathrm{norm}\!\left(\mathrm{boxed}(\hat{y}_i)\right)
=
\mathrm{norm}(g_i)
\right],
\]
where $N$ denotes the number of evaluated problems.

We do not assume that more recent or fine-tuned models necessarily achieve superior performance.
For each experimental configuration, models are first evaluated, and layer transplantation is performed from the empirically better-performing model to the weaker model.




\begin{table*}[t]
\centering
\small
\caption{
Best-performing transplantation settings between Phi-3 and Phi-3.5 across decoding token lengths.
}
\resizebox{\textwidth}{!}{
\begin{tabular}{c c c c c c c c c c}
\hline
Model Pair 
& Tokens
& Dir.
& Acc$_{\text{target}}$ (\%)
& Acc$_{\text{source}}$ (\%)
& Acc$_{\text{after}}$ (\%)
& Best $K$
& TIR
& Recovery (\%)
& Intervention Type \\
\hline

\multirow{5}{*}{Phi-3 (A) $\leftrightarrow$ Phi-3.5 (B)}
& 32
& B$\rightarrow$A
& 6.0
& 8.0
& 12.0
& 16
& 2.00
& 300.0
& FFN + Attention \\

& 64
& A$\rightarrow$B
& 16.0
& 20.0
& 20.0
& 16
& 1.25
& 100.0
& FFN + Attention \\

& 128
& A$\rightarrow$B
& 16.0
& 30.0
& 26.0
& 64
& 1.625
& 71.4
& FFN + Attention \\

& 256
& A$\rightarrow$B
& 34.0
& 34.0
& 36.0
& 8
& 1.059
& --
& FFN + Attention \\

& 512
& B$\rightarrow$A
& 36.0
& 40.0
& 40.0
& 8
& 1.111
& 100.0
& FFN + Attention \\
\hline
\end{tabular}
}
\label{phi3_phi35}
\end{table*}

\subsection{Results and Analysis}

In this section, we examine whether the proposed transplant technique can transfer task-relevant behavior between language models.
Furthermore, we investigate transplantation performance under different experimental settings, providing insights into how task-specific functionality is localized within model components.

First, we investigate how transplantation performance varies with respect to the number of selected layers $K$.


\begin{table}[t]
\centering
\small
\setlength{\tabcolsep}{6pt}
\caption{
Intervention composition as a function of transplantation scale $K$ for Phi-3 $\leftrightarrow$ Phi-3.5.
}
\renewcommand{\arraystretch}{1.0}
\resizebox{0.9\columnwidth}{!}{
\begin{tabular}{c c c c}
\hline
\textbf{$K$} & \textbf{FFN count} & \textbf{Attn count} & \textbf{Attn ratio} \\
\hline
8   & 7  & 1  & 0.125 \\
16  & 12 & 4  & 0.250 \\
32  & 14 & 18 & 0.563 \\
64  & 25 & 39 & 0.609 \\
128 & 64 & 64 & 0.500 \\
\hline
\end{tabular}
}
\label{composition_by_K_intra}
\end{table}

\begin{figure}[]
    \centering
    \includegraphics[width=0.46\textwidth]{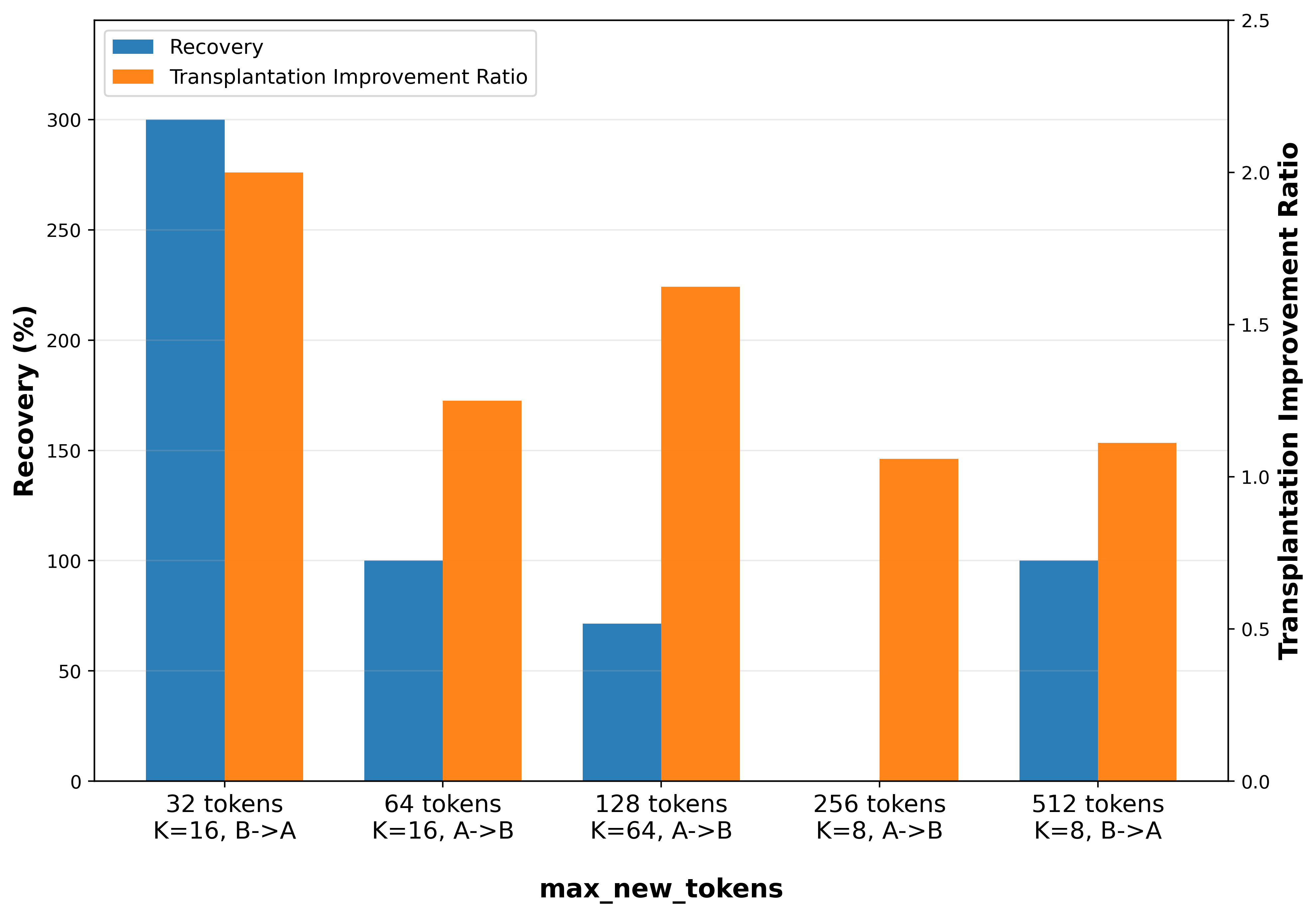}
    \caption{The TIR and Recovery of Phi-3 $\leftrightarrow$ Phi-3.5 as a function of the number of transplanted modules K for different decoding lengths with \texttt{max\_new\_tokens}.}
    \label{output_intra}
\end{figure}

For each decoding length, modules are transferred from the better-performing model to the weaker one, isolating capability transfer through structural intervention alone.

To clarify the notation in the experimental analysis, each term is defined as follows:
\begin{itemize}
    \item Tokens: The maximum decoding length used during evaluation, which determines the generation budget.
    \item Dir: the transplantation direction from the source to the target.
    \item Best $K$: The number of transplanted modules that maximizes Acc for the given decoding length.
    \item Transplantation Improvement Ratio (TIR): A target-referenced metric newly introduced in this paper to quantify the normalized improvement ratio of a transplanted model over the target baseline, defined as
    \begin{equation}
    \text{TIR} =
    \frac{\text{Acc}_{\mathrm{after}}}
    {\text{Acc}_{\mathrm{Target}}}.
    \end{equation}
    Positive TIR indicates a relative improvement over the target model.
    \item Recovery: We introduce a recovery metric that quantifies target-to-source gap closure after transplantation, measured as the fraction of the performance gap between the target and source baselines that is closed by the transplanted model.
    \begin{equation}
    \text{Recovery} = 100 \times 
    \frac{\text{Acc}_{\mathrm{after}} - \text{Acc}_{\mathrm{Target}}}
    {\text{Acc}_{\mathrm{Source}} - \text{Acc}_{\mathrm{Target}}}.
    \end{equation}
    A recovery value of $100$ indicates that the transplanted model reaches the source baseline, and a value above $100$ indicates that it surpasses the source baseline.
    \item Intervention Type: Indicates which module is configured when performing the transplant operation. 
\end{itemize}


\begin{table*}[]
\centering
\small
\caption{
Best-performing transplantation settings between Gemma-2-2B and Gemma-2-2B-IT across decoding token lengths.
}
\resizebox{\textwidth}{!}{
\begin{tabular}{c c c c c c c c c c}
\hline
\textbf{Model Pair} 
& \textbf{Tokens}
& \textbf{Dir.}
& \textbf{Acc$_\text{B}$ (\%)} 
& \textbf{Acc$_\text{A}$ (\%)} 
& \textbf{Acc$_\text{after}$ (\%)} 
& \textbf{Best $K$} 
& \textbf{TIR}
& \textbf{Recovery (\%)}
& \textbf{Intervention Type} \\
\hline

\multirow{5}{*}{Gemma-2-2B-IT (A) $\leftrightarrow$ Gemma-2-2B (B)}
& 32  
& B$\rightarrow$A
& 10.0
& 6.0  
& 10.0 
& 128  
& 1.667
& 100.0
& FFN + Attention  \\

& 64  
& A$\rightarrow$B
& 8.0 
& 8.0 
& 8.0 
& 32  
& 1.000
& --
& FFN + Attention  \\

& 128 
& A$\rightarrow$B
& 6.0 
& 22.0 
& 10.0 
& 64 
& 1.667
& 25.0
& FFN + Attention \\

& 256 
& A$\rightarrow$B
& 6.0 
& 36.0 
& 12.0 
& 64  
& 2.000
& 20.0
& FFN + Attention \\

& 512 
& A$\rightarrow$B
& 6.0 
& 30.0 
& 14.0 
& 64  
& 2.333
& 33.3
& FFN + Attention \\
\hline
\end{tabular}
}
\label{gemma_best_settings}
\end{table*}

\begin{table}[]
\centering
\small
\setlength{\tabcolsep}{6pt}
\caption{
Intervention composition as a function of transplantation scale $K$ for Gemma-2-2B-IT $\leftrightarrow$ Gemma-2-2B.
}
\renewcommand{\arraystretch}{1.0}
\resizebox{0.9\columnwidth}{!}{
\begin{tabular}{c c c c}
\hline
\textbf{$K$} & \textbf{FFN count} & \textbf{Attn count} & \textbf{Attn ratio} \\
\hline
8   & 0  & 8  & 1.000 \\
16  & 2  & 14 & 0.875 \\
32  & 2  & 30 & 0.938 \\
64  & 7  & 57 & 0.891 \\
128 & 39 & 89 & 0.695 \\
\hline
\end{tabular}
}
\label{composition_by_K_intra_gemma}
\end{table}

\subsubsection{Results between Model Generations}

Table~\ref{phi3_phi35} summarizes the best transplantation settings between Phi-3 and Phi-3.5 across decoding lengths, reporting post-transplant accuracy and the proposed metric, TIR.
Across token budgets, activation-selected transplantation yields measurable gains over the target baseline, while the optimal direction and transplantation scale depend on the decoding horizon.

Under short decoding, the best configuration favors transfer from the stronger model to the weaker model.
At 32 tokens, the best setting with $K{=}16$ improves accuracy from 6 to 12, corresponding to $\text{TIR}=2.0$ which means $\text{Acc}_{\mathrm{after}}$ is two times higher than $\text{Acc}_{\mathrm{Target}}$.
At 64 and 128 tokens, the best settings occur at moderate transplant scales and remain beneficial, with $\text{TIR}$ peaking at $1.25$ and $1.625$, respectively.
Recovery provides a complementary view based on the gap closure between the target and source baselines.
At 64 and 512 tokens, $\text{Recovery}=100.0$, showing complete recovery to the source level.  
At 32 tokens, the target-to-source gap is more than fully recovered, with $\text{Recovery}=300.0$, indicating that the transplanted model surpasses the source baseline. 
This over-recovery suggests that transplantation is not a mere functional copy of the source behavior, but a compositional recombination in which transplanted modules interact with the remaining target circuitry to yield a synergistic improvement beyond either baseline. 
At 128 tokens, $\text{Recovery}=71.4$, indicating partial recovery of the target-to-source gap despite a clear TIR gain. 
At 256 tokens, Recovery is not defined because the target and source accuracies are identical, making the denominator zero.
The intervention composition provides insight into the underlying mechanism.
As shown in Table~\ref{composition_by_K_intra}, small $K$ sets are Feedforward Network (FFN)-dominant, whereas attention modules become increasingly prevalent with larger values of $K$ grows.
This suggests that improvements at small scales are primarily driven by localized feed-forward transformations, while larger transplant scales increasingly engage attention pathways that mediate token-to-token interactions.

Increasing $K$ does not yield monotonic gains at longer decoding lengths.
Fig.~\ref{output_intra} shows that TIR often peaks at intermediate $K$ and can diminish at larger transplant scales.
This non-monotonic pattern indicates that beneficial transfer is selective in both scale and layer type, and that broader replacement does not guarantee additional gains, even under high structural compatibility.


\begin{figure}[]
    \centering
    \includegraphics[width=0.46\textwidth]{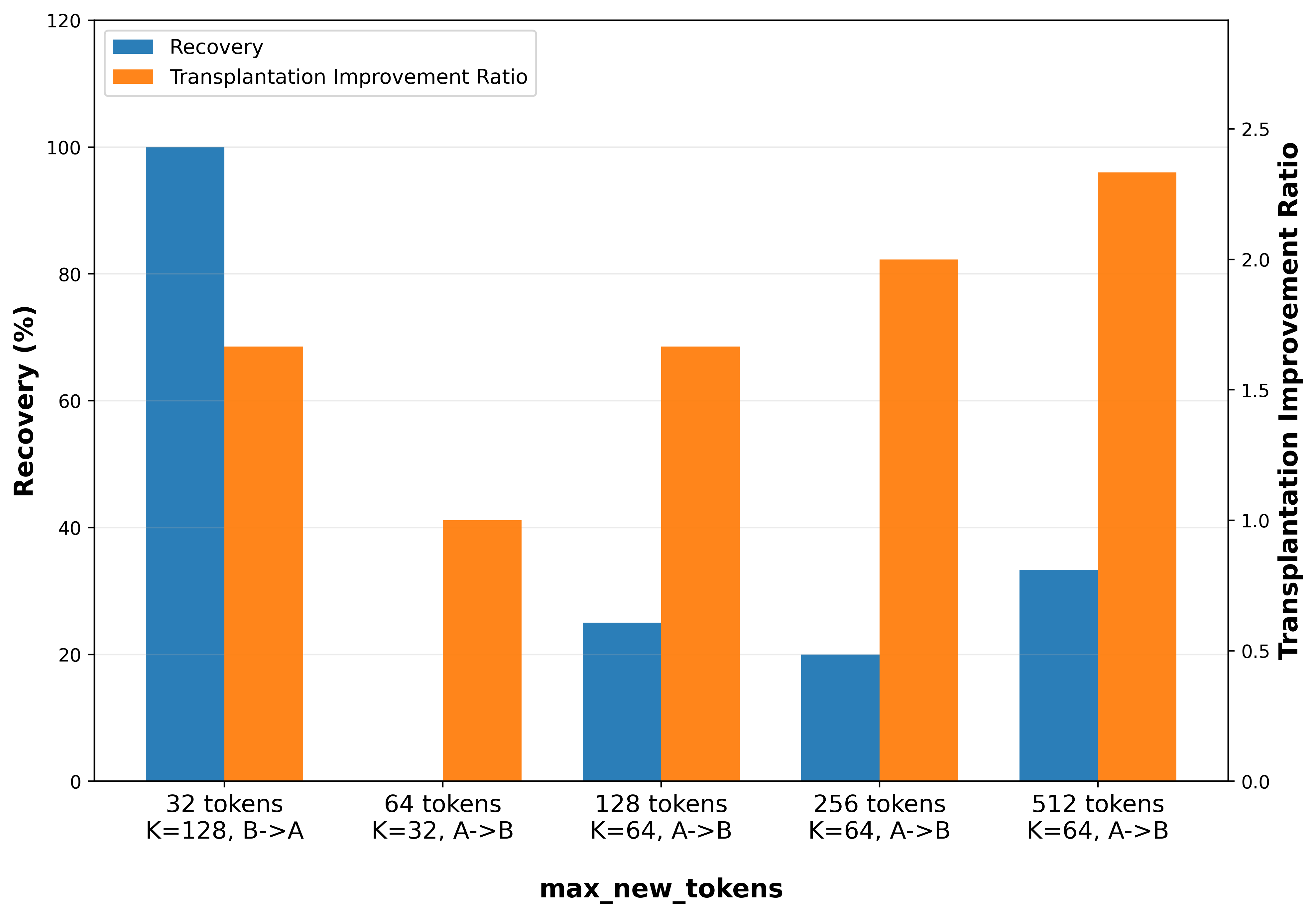}
    \caption{The TIR and Recovery of Gemma-2-2B-IT $\leftrightarrow$ Gemma-2-2B as a function of the number of transplanted modules $K$ for different decoding lengths with \texttt{max\_new\_tokens}.}
    \label{output_intra_gemma}
\end{figure}

\subsubsection{Results between Base and Tuned Models}

Table~\ref{gemma_best_settings} and Table~\ref{composition_by_K_intra_gemma} summarize transplantation between Gemma-2-2B model and Gemma-2-2B-IT model.
Because the two models share the same backbone architecture, this setting isolates the effect of post-training on transferability under high structural compatibility.

Table~\ref{gemma_best_settings} reports the best configuration at each decoding token length.
At 64 tokens, the two models exhibit the same baseline accuracy, and the best post-transplant accuracy remains unchanged, yielding $\text{TIR}=1.0$.
This indicates a saturation regime with little headroom for relative improvement under the given decoding constraint.
For moderate and long decoding token budgets, the optimal direction is predominantly \textsc{IT}$\rightarrow$\textsc{Base}, reflecting that the instruction-tuned model is substantially stronger.
The transplant composition further indicates that transfer in this setting involves broad coordination across attention pathways.
As shown in Table~\ref{composition_by_K_intra_gemma}, the optimal sets remain attention-dominant across $K$, with high attention ratios even at large intervention scales.

At 512 tokens, the best configuration is \textsc{Gemma-2-2B-IT}$\rightarrow$\textsc{Gemma-2-2B} with $K{=}64$, as shown in Fig.~\ref{output_intra_gemma}.
Under the proposed TIR metric, this corresponds to $\text{TIR}=2.333$, indicating a large relative gain over the target baseline at long decoding lengths.
Recovery provides a complementary gap-based view between the target and source baselines. 
At 32 tokens, $\text{Recovery}=100.0$, showing complete recovery to the source level. 
At 128, 256, and 512 tokens, Recovery remains partial at 25.0, 20.0, and 33.3, respectively, indicating that transplantation improves the target baseline while not fully closing the target-to-source gap under longer decoding horizons. 
At 64 tokens, Recovery is not defined because the target and source accuracies are identical, making the denominator zero.
Fig.~\ref{output_intra_gemma} shows that increasing $K$ does not yield monotonic gains and that TIR peaks at intermediate $K$ before diminishing at larger transplant scales for longer decoding lengths.
This is consistent with instruction tuning, which can induce broad changes in generation behavior and token-to-token interaction patterns beyond task-specific computation \cite{team2024gemma}.

The component differences in transplantation across $K$ reported in Table~\ref{composition_by_K_intra} and Table~\ref{composition_by_K_intra_gemma} arise from how activation mismatches distribute across modules for each model pair.
Phi-3 and Phi-3.5 form a cross-generation setting with differences in architecture and training, whereas Gemma-2-2B and Gemma-2-2B-IT primarily differ in the behavioral policies induced by instruction tuning.
Accordingly, in the Phi setting, performance-related activation differences are concentrated in localized representation transforms, and the largest activation mismatches at small $K$ tend to appear in FFN projections.
As a result, activation-based selection preferentially yields FFN-dominant intervention sets for small $K$, while attention modules are gradually included as $K$ increases.

In the Gemma base versus instruction-tuned setting, generation policies such as output format, response length, termination propensity, and conversational alignment differ and are more directly expressed through attention pathways that mediate token-to-token interactions.
Consequently, activation-based selection consistently yields attention-dominant intervention sets across $K$.

\subsection{Discussion}
Our experiments show that transplantation induces a localized functional shift: the transplanted modules move the target model responses toward those of the source model, while the remaining untransplanted modules largely preserve their original characteristics.
This supports a modular view in which task-relevant computation is concentrated in a sparse subset of internal modules rather than being uniformly distributed across the network.

An important implication is that effective transfer is selective not only in scale $K$, but also in model type and decoding token length.
In the Phi setting, the largest gains over the target baseline are achieved at small to moderate transplant scales, and the optimal transplant composition shifts as the decoding horizon increases.
In the Gemma base versus instruction-tuned setting, transplantation consistently improves the target model at long decoding lengths, while gains can saturate when the two models have similar baselines under a constrained token budget.

Overall, these results indicate that transfer is most effective when we transplant modules whose activations best match the target task under the given decoding length.
These findings motivate future work on identifying functionally equivalent substructures beyond strictly aligned architectures.

\section{Conclusion}

In this work, we provided experimental evidence of portability between language models for performance improvement.
We demonstrated immediate functional changes by directly transplanting a small subset of internal modules.
Using the TIR metric, we show that transplantation based on activation-based analysis can partially recover domain-specific features without retraining or parameter optimization.
Complementing TIR, the Recovery metric quantifies how much of the target-to-source performance gap is closed after transplantation, enabling a normalized assessment of transfer effectiveness across model pairs and decoding horizons.

Our results indicate that transferable computation is localized within specific layers, revealing an implicit modular structure inside pretrained language models.
These findings offer insight into how internal representations are organized and suggest that localized parameter replacement may serve as a viable mechanism for model editing.

We plan to extend transplantation to heterogeneous model pairs with architectural mismatches by developing compatibility mappings between non-identical modules.
Finally, we will broaden the evaluation beyond mathematical reasoning to diverse domains and tasks to better characterize when and how localized module replacement yields reliable capability transfer.

\section*{Acknowledgements}
This work was supported by the Korea Institute of Energy Technology Evaluation and Planning(KETEP) and the Ministry of Climate, Energy \& Environment(MCEE) of the Republic of Korea (RS-2022-KP002860), and also supported by the MSIT (Ministry of Science and ICT), Korea, under the ITRC (Information Technology Research Center) support program (IITP-2025-RS-2021-II211835) supervised by the IITP (Institute of Information \& Communications Technology Planning \& Evaluation).



\bibliographystyle{cas-model2-names}

\bibliography{cas-refs}


\end{document}